# FengWu-W2S: A deep learning model for seamless weather-to-subseasonal forecast of global atmosphere

**Authors:** Fenghua Ling[1†], Kang Chen[1†], Jiye Wu[2†], Tao Han[1], Jing-Jia Luo[2*], Wanli Ouyang[1], Lei Bai[1*]
**Affiliations:**
[1]Shanghai AI Laboratory, Shanghai, China
[2]Institute for Climate and Application Research (ICAR)/ /School of Future Technology /CIC-FEMD, Nanjing University of Information Science and Technology, Nanjing, China

†Equal Contributions,
*Corresponding to baisanshi@gmail.com; jjluo@nuist.edu.cn

**Seamless forecasting that produces warning information at continuum timescales based on only one system is a long-standing pursuit for weather-climate service. While the rapid advancement of deep learning has induced revolutionary changes in classical forecasting field, current efforts are still focused on building separate AI models for weather and climate forecasts. To explore the seamless forecasting ability based on one AI model, we propose FengWu-Weather to Subseasonal (FengWu-W2S), which builds on the FengWu global weather forecast model and incorporates an ocean-atmosphere-land coupling structure along with a diverse perturbation strategy. FengWu-W2S can generate 6-hourly atmosphere forecasts extending up to 42 days through an autoregressive and seamless manner. Our hindcast results demonstrate that FengWu-W2S reliably predicts atmospheric conditions out to 3-6 weeks ahead, enhancing predictive capabilities for global surface air temperature, precipitation, geopotential height and intraseasonal signals such as the Madden-Julian Oscillation (MJO) and North Atlantic Oscillation (NAO). Moreover, our ablation experiments on forecast error growth from daily to seasonal timescales reveal potential pathways for developing AI-based integrated system for seamless weather-climate forecasting in the future.**

Over the past few decades, the increasing demand for accurate weather and climate forecasts across various timescales has led to significant efforts in developing seamless prediction systems based on dynamical models[1–4]. These systems aim to provide decision-makers with daily, subseasonal, seasonal, interannual, and decadal forecasts. However, this traditional approach costs huge computing resources, facing substantial challenges and slow progress. In recent years, the rapid development of AI-based models has transformed global weather forecasting, offering innovative methodologies and perspectives[5–10]. Compared to classical dynamical models, AI models are increasingly recognized for their superior prediction accuracy[11–15]. Given their ability to generate forecasts quickly, efficiently, and economically, a question is that: can the current AI-based medium-range weather forecast models can be extended to provide accurate predictions over longer timescales in a seamless manner?

While many existing AI models attempt to provide forecasts across different time scales by reducing time resolution and restructuring their frameworks[16–19], these approaches often fail to accurately capture the intricate interactions among various time scales and do not represent a true integrated forecast system that links weather and climate seamlessly. In contrast, traditional dynamical models have made strides toward seamless predictions through the development of coupled dynamical models[20–22] and the implementation of ensemble forecast strategies[23–25]. These methodologies have proven effective in capturing the multi-scale interactions and improving forecast skills. This leads to another critical question: can these traditional techniques be employed to current AI-based weather models, so that they can also capture the multi-scale interactions and produce extended and improved forecasts seamlessly from weather to climate?

To address these questions, we propose FengWu-W2S, an extension of the medium-range AI weather model (FengWu[7]). Based on the assessment of hindcasts during the past twenty years, we find FengWu-W2S can accurately predict atmospheric conditions up to 3 to 6 weeks in advance, and compared to traditional numerical models, it also significantly enhances the forecasts of intraseasonal phenomena such as the Madden-Julian Oscillation (MJO) and the North Atlantic Oscillation (NAO). Furthermore, we explored possible reasons for the higher accuracy of the FengWu-W2S forecasts. Our results demonstrate that the methods employed in dynamical model can also be applied to current AI models to improve and generate weather-subseasonal forecasts. This suggests that the AI models can also provide a feasible pathway for developing an "all-in-one seamless forecasting system".

## Result

FengWu-W2S adopts the same prediction framework as FengWu, using a 6-hourly initial field to generate forecasts every 6 hours for the next 42 days through autoregression (i.e., 168 iterations). To enhance the representation of interactions between the atmosphere, ocean, and land, we add more relevant variables and sub-seasonal predictability sources (see Methods, Fig. S1). In the model building process, we aim to better simulate ocean-atmosphere and land-atmosphere coupling, with a focus on the inter-exchange of surface variables. Unlike FengWu weather forecast model, which aggregates features from all surface variables, we design a feature extraction module to explicitly represent the energy and mass exchanges among the three components through inter-module interactions (see section "*Land-Ocean-Atmosphere coupling in AI model*"). Additionally, we introduce the multi-member ensemble strategy for providing probabilistic and better forecasts, including initial condition perturbations and adaptive controllable model perturbations. They are adopted to estimate the uncertainty in initial conditions and AI model parameters, respectively, and can better capture sub-grid physical processes[26] and extreme events[27] (see section "*Ensemble forecast strategy*").

### *Weather-subseasonal prediction skills of FengWu-W2S*

Firstly, we assessed the forecast skill (temporal anomaly correlation coefficient, ACC) of weekly anomalies of the most important variables—air temperature at 2 meter (T2m), total precipitation (TP), geopotential height at 500 hPa (Z500), and outgoing longwave radiation (OLR)—across different lead weeks (Fig. 1). Interestingly, while awaiting the generation of multi-year hindcasts, we found that FengWu-W2S exhibits unexpectedly high forecast skill when using observed climatology, especially for precipitation (Fig. 1B). Besides this, regardless of whether observed (FengWu-W2S$_o$) or model climatology (FengWu-W2S$_m$) is used, the predictive skill of FengWu-W2S consistently surpasses that of traditional methods, including both dynamical and current AI models, especially for T2m and Z500, even with a significantly larger number of iterative forecast steps (Fig. 1).

Furthermore, we presented the spatial distribution of deterministic and probabilistic skills in predicting precipitation and 2m air temperature anomalies at lead times of 3-4 and 5-6 weeks (Fig. 2). The ACC skill of 2m air temperature prediction indicates that forecasts at lead times of 3–4 weeks can correctly capture the temperature phase changes across most regions of the globe. However, at lead times of 5-6 weeks, negative forecast skills emerge in mid to high-latitude areas, such as coastal East Asia and the North American Atlantic, regardless of whether FengWu-W2S$_o$ or FengWu-W2S$_m$ (Fig. S2). In contrast, pronounced climate signals, such as El Niño-Southern Oscillation and Indian Ocean Dipole in the equatorial oceans, continue to display relatively strong predictive skills, indicating that FengWu-W2S effectively captures long-lasting air-sea interactions. Nevertheless, FengWu-W2S struggles to accurately predict extreme heat wave events over certain areas, such as in the eastern equatorial Pacific, where the Brier skill score calculated for 90th percentile (BSS) is significantly lower than the climatology forecast at lead time of 5-6 weeks. Additionally, tercile ranked probability skill score (RPSS) indicates that accurately forecasting temperature changes over land areas remains challenging at 5-6 weeks lead even with probability forecasts, and FengWu-W2S predictions generally fall below climatology forecasts.

Regarding precipitation forecast, while most regions demonstrate positive predictive skills, the overall skills are relatively low. Regions with frequent tropical precipitation, particularly the equatorial areas, show higher predictive skills. Interestingly, compared to precipitation forecast skill (i.e., ACC) calculated based on the model's climatology, the skill calculated based on the observed climatology displays relatively higher TCC outside the tropics (cf. Fig. 2 and Fig. S2). Regarding extreme flood events and tercile-based precipitation probability forecasts, a "dry-mask" was applied to areas with no rain (i.e., biweekly cumulative precipitation is less than 1 mm/2week). FengWu-W2S accurately predicts extreme events in tropical convective areas with heavy rainfall and effectively distinguishes between dry, normal, and wet conditions relative to climatology forecasts. However, in regions experiencing persistent drought, such as the eastern equatorial Pacific, the skill score is worse than a coin toss.

***Prediction skill of major intra-seasonal signals***

Previous studies suggest that weather and subseasonal predictability is influenced by climate variation modes operating on intra-seasonal timescales, particularly the Madden–Julian Oscillation[28] (MJO) and the North Atlantic Oscillation[29] (NAO). When these modes are active and strong, model forecasts often show enhanced skill. Therefore, it is essential to assess the forecast performance of these predictability sources in FengWu-W2S.

**MJO**, the most dominant mode of tropical atmosphere variability on intra-seasonal time scales, characterized by large-scale convective anomalies and associated circulation anomalies that propagate eastward from the tropical Indian Ocean to the Pacific, significantly influencing global weather patterns[30–32]. The FengWu-W2S model demonstrates impressive performance in predicting the MJO, achieving a skillful prediction (i.e., TCC >0.5) at lead time of up to 37 days, longer than 32 days in ECMWF (Fig. 3c). Correspondingly, the eastward propagation of MJO convection is effectively captured at the lead time of 30 days, albeit with reduced intensity (cf. Fig. 3a and 3b). The composited phase diagrams of the FengWu-W2S starting from eight different MJO phases align well with observed counterparts, particularly at lead times shorter than three weeks. However, at longer lead times, the FengWu-W2S model slightly underestimates the MJO amplitude (Fig. 3d). Notably, one strong MJO event occurring in February through April in 2019 is accurately predicted at lead times of up to 40 days (Fig. 3e). It is also worth mentioning that the ACC skill of FengWu-$W2S_o$ is slightly better than that of FengWu-$W2S_m$, particularly at lead times beyond 20 days (Fig. 3c).

**NAO** is a pronounced mode of the extratropical atmospheric variability, with a profound influence on the weather and climate in the Northern Hemisphere, especially over Europe and eastern North America[33]. Operating on timescales from days to decades, it typically peaks during boreal winter[34,35]. To assess NAO prediction skill at subseasonal timescales, we focused on daily NAO index forecasts initialized from each day of December-February[36]. FengWu-$W2S_m$ exhibits comparable skill to the ECMWF, producing skillful forecasts with ACC above 0.5 up to 20 days lead (Fig. 4A). In particular, FengWu-$W2S_o$ can even provide skillful NAO forecast out to 28 days lead with a stable performance at lead times of 15-28 days. Compared to ECMWF, FengWu-$W2S_o$ demonstrates equivalent skills at lead times of 1-15 days but higher skills beyond 18 days (Fig. 4A). While forecast skills of weekly mean NAO are also comparable at 1-3 weeks lead, the skills of FengWu-$W2S_o$ again surpass those of ECMWF at lead times of 4-6 weeks (Fig. 4B). In addition, we also selected two typical NAO events in 2017/18 winter to assess the performance of FengWu-W2S on predicting the evolution of NAO (Fig. 4C). The results show that the model is able to predict the phase changes of NAO, including the negative NAO phase responsible for the cold wave at the end of February 2018. The ensemble forecast members capture the observed evolution reasonably well, particularly in the first 10 days. Note that FengWu-$W2S_o$ performance is again better than FengWu-$W2S_m$, suggesting the benefit of using observed climatology for anomaly calculation.

**Teleconnection patterns** are significant contributors to climate and weather anomalies over many regions worldwide. In this analysis, we evaluate the prediction skills of five major teleconnection patterns in Northern Hemisphere, including the Pacific-North American teleconnection (PNA), the East Atlantic teleconnection (EA), the West Pacific teleconnection (WP), and the Eurasian-Pacific teleconnection (EU) during boreal winter, as well as the Pacific-Japan teleconnection (P-J pattern) during boreal summer (Fig. 4D). The results show that FengWu-$W2S_o$ is able to skillfully forecast these important patterns at lead times of 1-2 weeks. However, its performance declines afterwards and produces low skills at 4-6 weeks lead. Among the five patterns, forecast skill of the PJ pattern is the highest, with good skills being extended to 4-5 weeks.

***Possible reasons for the improved performance of FengWu-W2S from weather to subseasonal timescales***

The community has begun to recognize that AI models often outperform traditional dynamical models, but with an increasing demand for improving the interpretability of AI models[11,18,37–40]. However, achieving interpretability that is consistent with physical law remains a great challenge. Therefore, we discuss whether AI models possess the capability to make predictions over longer time scales from the perspectives of error accumulation and predictive skill, rather than focusing on extracting physical interpretability.

We compared the error growth of FengWu and FengWu-W2S over a three-month forecast period, using a six-hourly temporal resolution with 360 autoregressive steps in total (Fig. 5A). Notably, FengWu-W2S exhibits a lower error growth rate at lead times between 10 and 42 days, reaching equilibrium sooner with reduced errors. While this characteristic may enhance its long-term forecasting ability, it is important to note that errors continue to grow with increasing iterations, albeit slowly. Unlike dynamical models, which rely on partial differential equations, AI models lack physical framework constraints. This absence can lead to more random errors rather than systematic ones during forecast iterations, which may explain why the climatology of the AI model display larger errors.

To understand why FengWu-W2S maintains a lower cumulative error than FengWu does, we conducted ablation experiments. Our analysis indicates that incorporating data from additional subsystems enhances predictive capability(FengWu-AOL), while explicit interaction modules among these subsystems improve data utilization (Fig. 5B, solid lines). Moreover, perturbation strategies significantly enhance the subseasonal predictive skill, even without the additional data input. While initial condition perturbations improve predictive skill at 2 to 3 weeks lead, their effectiveness decrease after the lead time of 4 weeks and often lead to decreased accuracy during the first week. Combining the model perturbations with initial perturbations can mitigate these issues and effectively enhance long-range predictive capability (see Fig. 5B, dashed lines).

Given that the error growth of FengWu-W2S stabilizes at lead times between 42 days and 3 months, we wonder whether the model possesses the capability for subseasonal-to-seasonal forecasting. To investigate this, we evaluated the model's performance in predicting monthly mean anomalies. The results indicate that even at 3-month lead, FengWu-W2S demonstrates a certain level of skill for the global T2m forecast (Fig. 5C). This suggests that, as long as more robust and reliable structures can be designed to optimize long-range forecast errors, AI models may have the potential to extend their forecast beyond the subseasonal timescale. In the future, all-in-one AI-based seamless forecast model would become a reality.

It is worth noting that the increase in TCC of FengWu-W2S$_o$ relative to FengWu-W2S$_m$ primarily originates from extratropical regions, while the tropical region shows little difference and even a slight decrease (Fig. 5D and Fig. S2). This discrepancy may be linked to faster error accumulation outside the tropics. This warrants further exploration of the underlying reasons.

**Conclusions**

In this study, we built FengWu-W2S, which enhances and extends FengWu performance by incorporating explicit interactions among multiple subsystems and employing various ensemble forecasting strategies. FengWu-W2S demonstrates that current AI models share similarities with traditional dynamical models, as both can extend predictive skill from medium-range to longer time scales by integrating model structure designs that align with physical principles, rather than simply reducing the resolution of time steps. In the future, we can draw on the techniques and methods developed for traditional models, refining them for use in AI models to improve the forecast skills. These approaches indicate a new way for development of seamless prediction or even all-in-one forecasting systems in the future.

However, the current FengWu-W2S model has several shortcomings, including its inability to simulate sudden stratospheric warming (SSW), lack of sea ice signals, and a coarse spatial resolution of only 1.4°. Enhancing the model's resolution and incorporating additional physical processes could improve its forecasting capabilities further at weather-subseasonal timescales. Furthermore, the current ensemble forecast skills can approach or even exceed the performance of IFS-ENS on medium-range weather forecasts through various ensemble strategy modifications (Fig. S3). In the future, developing more diverse and reliable perturbation strategies, such as breeding vectors[41] and conditional nonlinear optimal perturbations[42], could further enhance forecast accuracy.

Finally, we generated over 300 terabytes of hindcast data, incurring computational costs that far exceeded those of model training. Surprisingly, we found that using the AI model's own climatology does not help correct systematic biases for better skill, although this is usually true for evaluating traditional dynamical models forecast. This raises an interesting question. The AI model uses original data as regression targets, focusing on closely matching observed values rather than simulating atmospheric evolution using PDEs. Thus, the question of whether statistical models possess a climatology remains disputable. While we also conducte extensive numerical experiments, this remains an unsolved mystery, future studies will delve deeper into this issue. Future studies will delve deeper into this issue. If leveraging observed climatology can enhance forecast performance, we may eliminate the need for redundant hindcast data, significantly reducing overall computational costs.

## Methods

### *Data*

For training FengWu-W2S, we created datasets from the ERA5 dataset, which has a spatial resolution of 1.4° latitude/longitude with 128×256 grid points globally. The dataset spans from 1979-01-01 to 2016-12-31, and is downsampled from hourly data to six-hour intervals, except total precipitation for which we adopt 6-hourly accumulated values. For the skill assessment using testing data from 2017 to 2021, anomalies for all variables are defined as deviations from the climatology calculated over the 15-year period from 2002 to 2016.

To provide weather-subseasonal predictions of as many variables as we can and to simulate links between different variables, we employ a total of 78 variables. These include five upper-air atmospheric variables measured across 13 pressure levels (50, 100, 150, 200, 250, 300, 400, 500, 600, 700, 850, 925, and 1000 hPa), as well as 13 single-level variables. The 13-level atmospheric variables consist of geopotential (Z), temperature (T), zonal wind (U), meridional wind (V), and specific humidity (Q). And the single-level variables include air temperature at 2 meters high (T2m), total precipitation (TP), sea surface temperature (SST), outgoing longwave radiation (OLR), water volume in the soil layer from 0 to 7 cm (SWVL1), zonal wind at 10 meters and 100 meters high (U10, U100), meridional wind at 10 meters and 100 meters high (V10, V100), mean sea-level pressure (MSL), as well as significant wave height (SWH), mean wave direction (MWD) and mean wave period (MWP) on ocean surface. Table S1 presents a complete list of these variables along with their abbreviations.

### *Land-Ocean-Atmosphere coupling in AI model*

Inspired by traditional dynamical models, FengWu-W2S adopts the multi-modal architecture of FengWu, independently encoding different atmosphere and ocean variables. In FengWu-W2S, the variables are encoded separately to extract their high-dimensional information, which is then combined in a multi-modal fuser to simulate dynamic equations. The decoder subsequently restores the results to the spatial domain for each variable, yielding next-step predictions (see Fig. S1A). This independent architecture facilitates the design of explicit multi-subsystem interactions.

The most critical aspect of multi-subsystem coupling is the energy and mass exchange at the boundaries. Unlike FengWu, which combines all surface variables to extract high-dimensional features, we have designed a more physically interaction way to represent these exchanges during the encoding and decoding phases (see Fig. S1B). For instance, wind and pressure are encoded in the same feature space to reflect the process of wind-pressure adjustment, while temperature, precipitation, and sea surface temperature are encoded together to represent exchanges of surface heat and water vapor flux. Additionally, interactions among lower-layer variable in multi-modal fuser can model various physical processes, such as wind-wave interactions and the influence of tropical convection on temperature and precipitation. In addition to these surface variables adapt to one another, they will involve in deeper information interactions with the upper-layer circulation.

***Ensemble forecast strategy***

Considering the large uncertainties in weather-subseasonal forecasts, deterministic forecasts produced by only one member often cannot provide sufficient information with relatively lower skill. In contrast, ensemble forecasts with a large number of members can account for the uncertainties and provide probabilistic warning information with better skill and better performance for predicting extreme events[24,25,36]. In order to be more in line with the typical practice adopted in traditional models, we design two perturbation schemes: one with perturbed initial conditions and the other with AI model parameter perturbations (Fig. S1C). We generated a hindcast dataset from 2002 to 2022 with 30 ensemble members, and an extended version of 100 members for the period from 2017 to 2022.

**Initial Condition Perturbation** is a useful strategy in the classical ensemble forecasting. Specifically, this is to superimpose initial disturbances on the initial analysis field to form ensemble forecasts when there is a large uncertainty in the initial conditions[26]. Adding different initial disturbances can produce different forecasts, generating individual member forecasts and thus forming the ensemble forecast system. In our study, we utilize Perlin noise method to generate the normalized initial conditions, as implemented in Pangu-Weather, rather than relying on initial perturbations from the ECMWF Ensemble Data Assimilation (EDA) system[11,26]. This approach is preferred because the amplitude of the perturbations from the EDA system may be too small to be effective after normalization. In addition, we have also tried to use the diffusion model for data assimilation to add initial perturbations, but found that the spread of initial condition is still relatively small, and the consumption of computing resources is increased several times.

**Adaptive Controllable model Perturbation** is an ensemble forecast strategy we use to simulate the perturbations of physical processes in dynamical models. It can be used to estimate the uncertainty of the model and effectively captures physical processes that are parameterized in operational NWP models[26]. As shown in Fig. S1A, current deep learning models involve multi-level encoding and decoding, leading to the extraction of features from low-level to high-level spaces. For instance, taking the geopotential height at 500 hPa level as an example, the upper layer may capture high-frequency signal distribution patterns, the second layer after downsampled may represent major planetary systems such as the Subtropical Highs, and bottom layers may focus on global distribution patterns. Given the different features operate at various spatial scales, the applied perturbations should also vary accordingly. Therefore, we implement adaptive multi-layer perturbations by introducing multi-layer variational autoencoders (VAEs) during the fine-tuning process. This approach ensures that perturbations are tailored to the specific features extracted at each layer and for each variable.

The weather-subseasonal forecasts are short compared to climate prediction and hence model biases may play a less important role. Uncertainty in the predictions of intraseasonal signals such as the MJO and variations of soil moisture (SWVL1) may primarily arise from uncertainties in initial conditions. Beside this, our experiments showed that perturbing the model structure of certain variables, such as T2m and SST, significantly reduces predictive ability (figure not shown). To address this, we designed a module with a switch to toggle the perturbation of specific variables on and off, enhancing the simulation of physical process perturbations and improving forecast robustness. In the future, we aim to develop this switch into an automated system that learns and adjusts perturbations based on training results.

### *Training details*

FengWu-W2S is implemented using the PyTorch framework and utilizes 8 Nvidia A100 GPUs for two-stage training, with a total batch size of 32. In the first stage, training is conducted for 50 epochs using a probability loss[7]. In the second stage, building on the weights from the first stage, a multi-layer VAE module is introduced to apply model perturbation, employing a multi-layer Kullback–Leibler divergence loss ($KL_{loss}$) to regulate the amplitude of the perturbation. This stage is fine-tuned for an additional 40 epochs. The optimization is carried out using the AdamW optimizer with the following parameters: β1 = 0.9, β2 = 0.9, and an initial learning rate of $5 \times 10^{-4}$. The second stage loss function defined as follows:

$$Loss = P_{loss}(\hat{Y}, Y) + \lambda(\sum_{i=1}^{n} KL_{loss}(x_i, \varepsilon)) \quad (1)$$

where $P_{loss}$ represents the probability loss similar to that described in FengWu (see Chen et al., 2023[7]). The $\hat{Y}, Y$ and $x_i$ correspond to model output, ground truth and features from different layer of different encoders, respectively. The parameter $\lambda$ refers to a tune-able coefficient balancing $KL_{loss}$ and $P_{loss}$, is set to $1 \times 10^{-4}$in this study. $\varepsilon$ represents Gaussian noise with 0-1 distribution.

### *Evaluation metrics*

Based on the prediction, we compute some metrics, i.e., the latitude-weighted ACC, BSS[43], and RPSS[44] defined as follows:

$$ACC(c,t,i,j) = \frac{\sum_{t_0 \in D} f'_{c,t_0+t,i,j}\, o'_{c,t_0+t,i,j}}{\sqrt{\sum_{t_0 \in D} (f'_{c,t_0+t,i,j})^2 \sum_{t_0 \in D} (o'_{c,t_0+t,i,j})^2}} \quad (2)$$

where $t_0$ represents the initialization time in the testing dataset. The indices $c, i$ and $j$ correspond to variables, latitude, and longitude coordinates, respectively. $t$ refers to the lead time steps. $f'$ and $o'$ are the anomaly of the forecast and observation, respectively. Unlike previous evaluations, the FengWu-W2S assessment utilized two climatologies, calculated from both FengWu-W2S hindcasts and observation, respectively. These averages span a 15-year period from 2002 to 2016.

To compute the RPSS, $K$ categorical forecasts are first defined (tercile). Then let $P_f(i)$ be the probabilistic forecast of the event occurring in category $i$; let $P_c(i)$ be the climatological probability of the event falling in category $i$. We bin the verification such that $P_o(i) = 1$ if the event was observed to be in category i, and $P_o(i) = 0$ otherwise. The $k^{\text{th}}$ components of the cumulative forecast, climatological, and observational distributions $F_{fk}$, $F_{ck}$ and $F_{ok}$ are evaluated for each of the K categories as $F_{fk} = \sum_{i=1}^{k} P_f(i)$, probability scores for the forecast ($RPS_f$) and climatology ($RPS_c$) are computed as

$$RPS_f = \sum_{k=1}^{K} \left(F_{fk} - F_{ok}\right)^2 = \sum_{k=1}^{K} \left(\sum_{i=1}^{k} P_f(i) - \sum_{i=1}^{k} P_o(i)\right)^2 \quad (3)$$

$$RPS_c = \sum_{k=1}^{K} (F_{ck} - F_{ok})^2 = \sum_{k=1}^{K} \left(\sum_{i=1}^{k} P_c(i) - \sum_{i=1}^{k} P_o(i)\right)^2 \quad (4)$$

Use angle brackets $\langle\cdot\rangle$ to denote the average of the scores over a given number of forecast-observation pairs, the RPSS are computed as

$$RPSS = 1 - \frac{\langle RPS_f \rangle}{\langle RPS_c \rangle} \quad (5)$$

In the special case of forecasts with exactly two categories (90$^{\text{th}}$ climatological percentiles), the RPS becomes the well-known BS and the RPSS becomes the BSS, respectively:

$$BSS = 1 - \frac{\langle BS_f \rangle}{\langle BS_c \rangle} \quad (6)$$

**Data and materials availability:**

We downloaded a subset of the daily statistics from the ERA5 hourly data from the official website of Copernicus Climate Data (CDS) at https://cds.climate.copernicus.eu/cdsapp#!/software/app-c3s-daily-era5-statistics. The ECMWF S2S data were obtained from https://apps.ecmwf.int/datasets/data/s2s/.

**Acknowledgments:** J.-J L is supported by National Natural Science Foundation of China (Grant No. 42088101), National Key R\&D Program of China (Grant No. 2020YFA0608000).

**Figure and Tables:**

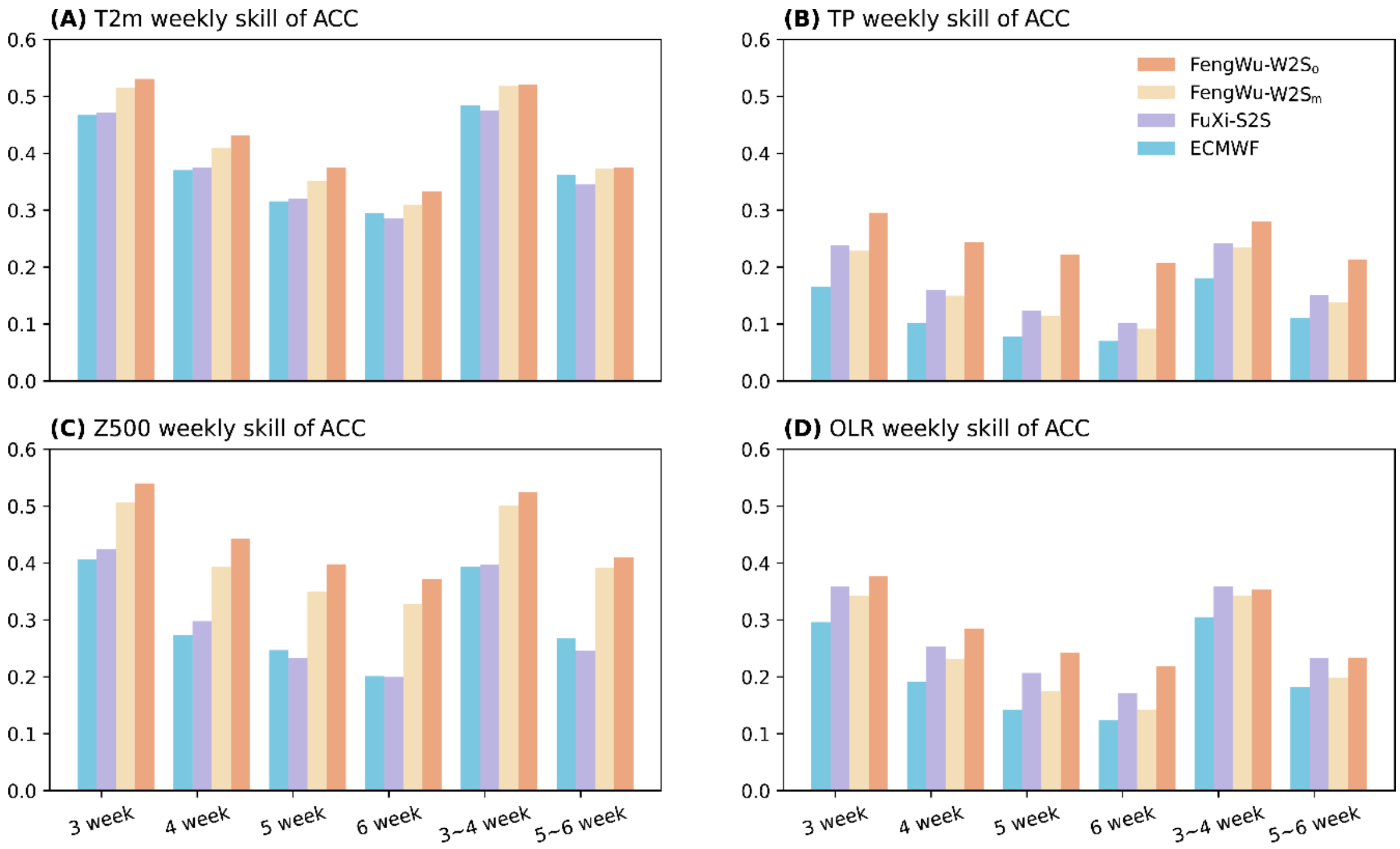

**Figure 1 Correlation skills of subseasonal predictions among different models.** (A) Predictive skill of 2m air temperature (T2m) anomalies as a function of lead week based on ECMWF (blue bar), Fuxi-S2S (purple bar) and FengWu-W2S forecasts. The skills of FengWu-W2S are assessed based on the anomalies calculated relative to the observed climatology (orange bar) and FengWu-W2S's hindcast climatology (yellow bar), respectively. The prediction skill is validated for the period of 2017–2021. (B, C, D) As in (A), but for the prediction skills of the anomalies of total precipitation (TP), T2m, and geopotential height at 500 hPa (Z500), respectively.

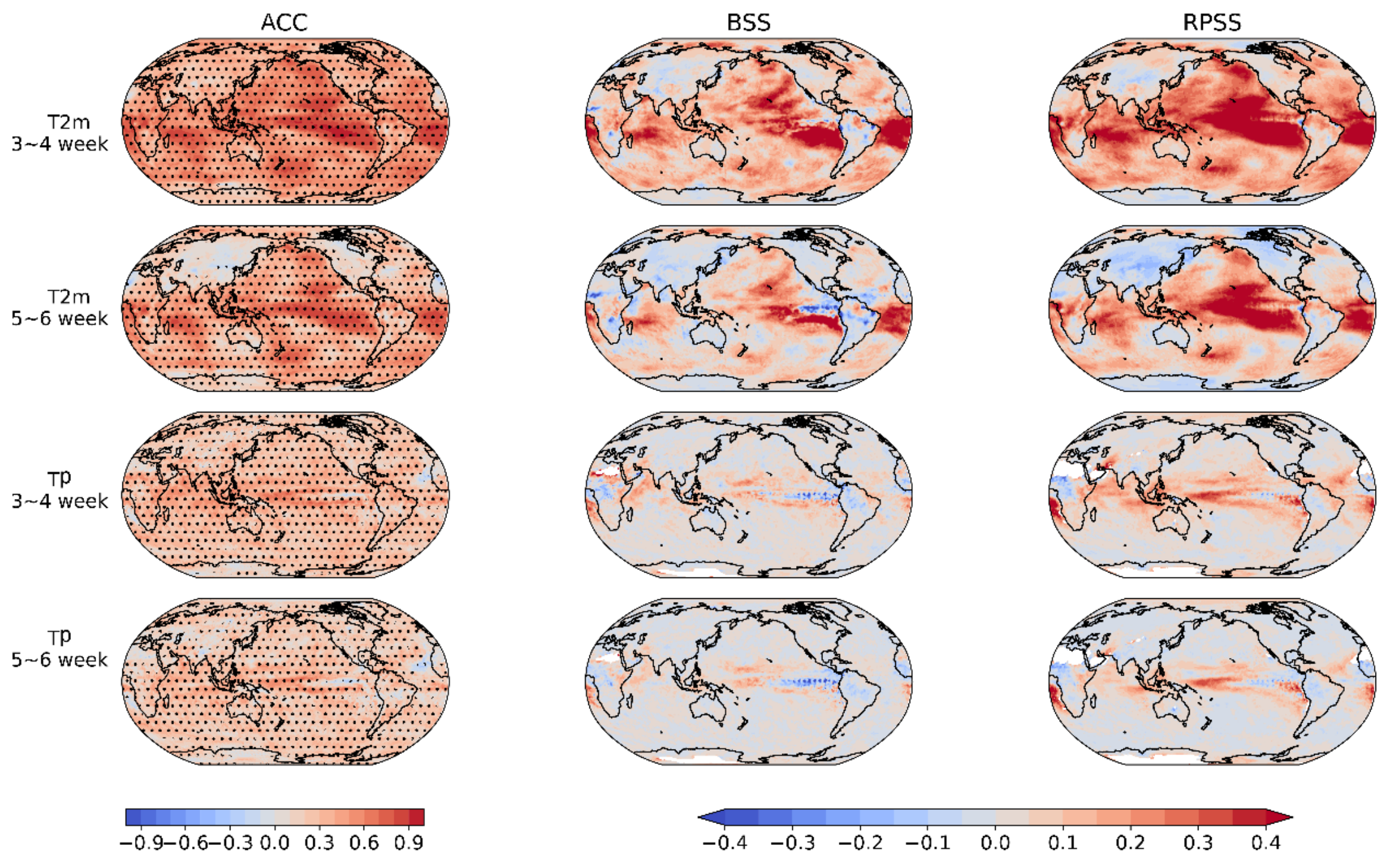

**Figure 2 Spatial distribution of prediction skills based on FengWu-W2S.** The spatial map of biweekly forecast skill in predicting T2m (1st and 2nd row) and TP (3rd and 4th row) anomalies based on FengWu-W2S model. The evaluation metrics include the deterministic skill, i.e. temporal anomaly correlation coefficient (TCC, first column), and probabilistic forecast skills, i.e., brier skill score (BSS, second column) and ranked probability skill score (RPSS, third column). The prediction skill is validated for the period of 2017–2021.

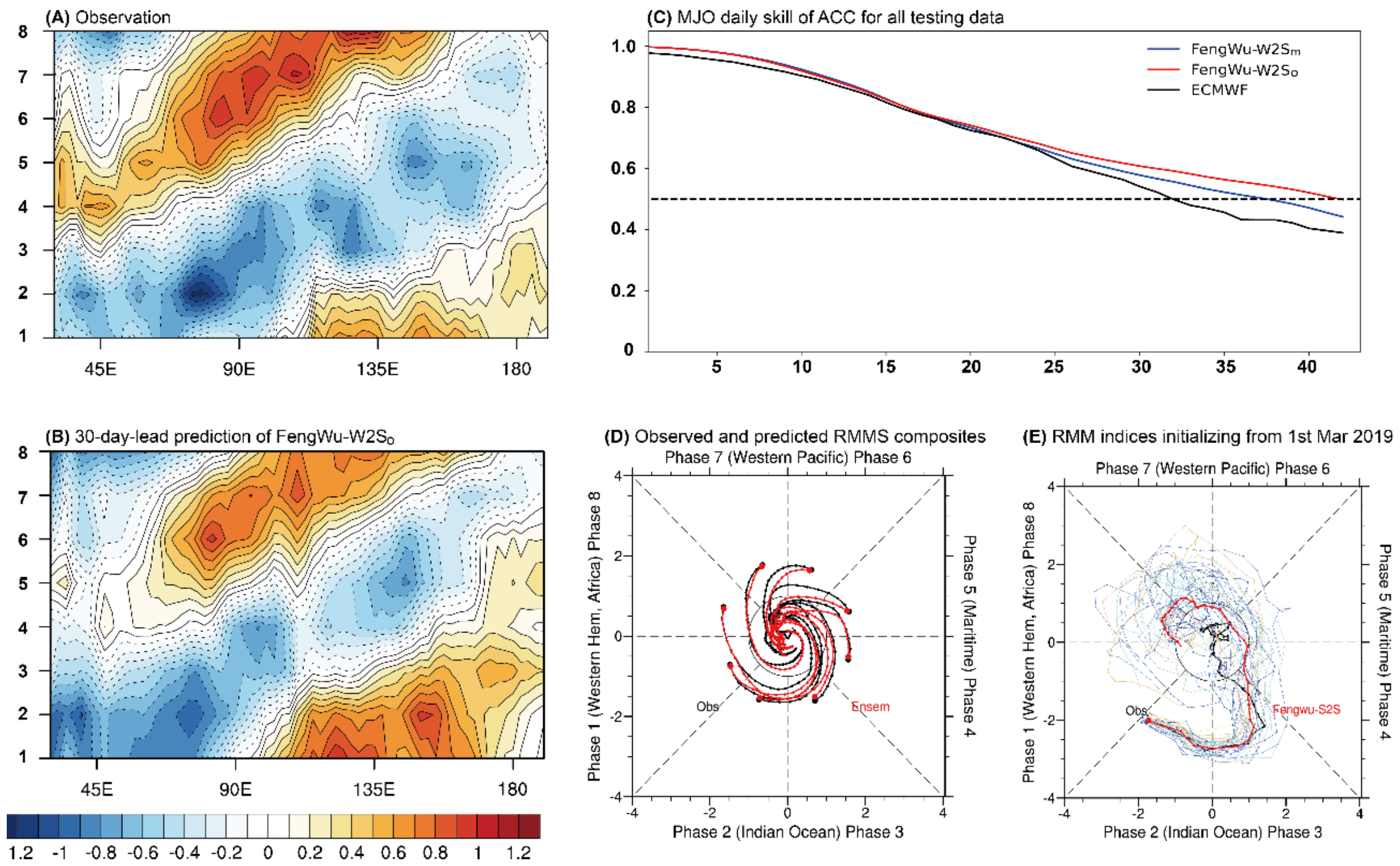

**Figure 3 The MJO index forecast skills of FengWu-W2S.** Hovmöller diagrams of normalized intraseasonal OLR anomalies (units: $W \cdot m^{-2}$) averaged over 15°N-15°S based on the composite of 8 phases in the (A) observation and (B) 30-day-lead prediction of FengWu-W2S. (C) TCC skill in predicting Realtime Multivariate MJO (RMM) index with the anomalies being calculated relative to the climatology of observation (blue line) and model hindcast (red line), respectively. (D) Phase-space diagram of the composited RMM index initializing from strong MJO (amplitude >1 ) at eight phases based on the observation and FengWu-W2S predictions up to 40 days lead. (E) Phase-space diagram of the observed RMM index (black line) and FengWu-W2S ensemble mean (red line) and individual member (other color lines) forecasts initialized from 1st Mar 2019. Note that the anomalies used in a, b, d, e are calculated using the observed climatology.

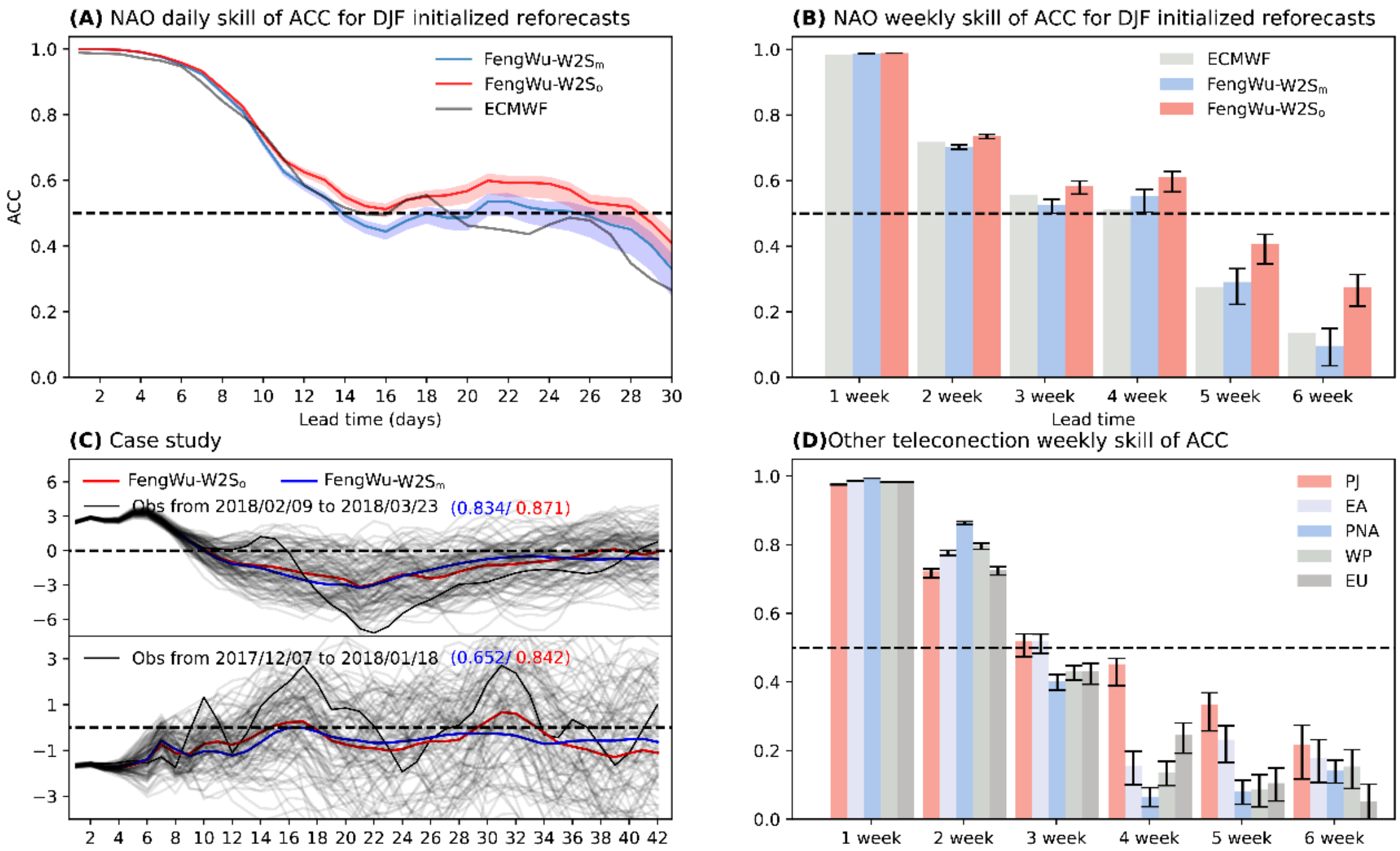

**Figure 4 Forecast skills of the NAO index and five teleconnection indices based on FengWu-W2S. (A)** Predictive skill of the NAO daily index as a function of the lead time based on the hindcast of ECMWF (black line), FengWu-W2S relative to observed climatology (red line) and model's own climatology (blue line). (B) As in (A), but for NAO weekly index. (C) FengWu-W2S forecasts of two NAO events at 1-42 days lead. The gray lines represent the predictions of different members. The blue (red) lines display the forecast anomalies relative to the model's (observed) climatology, and the values in blue (red) represent their TCC skills. (D) Weekly forecast skill of five major teleconnection patterns based on FengWu-W2S$_o$.

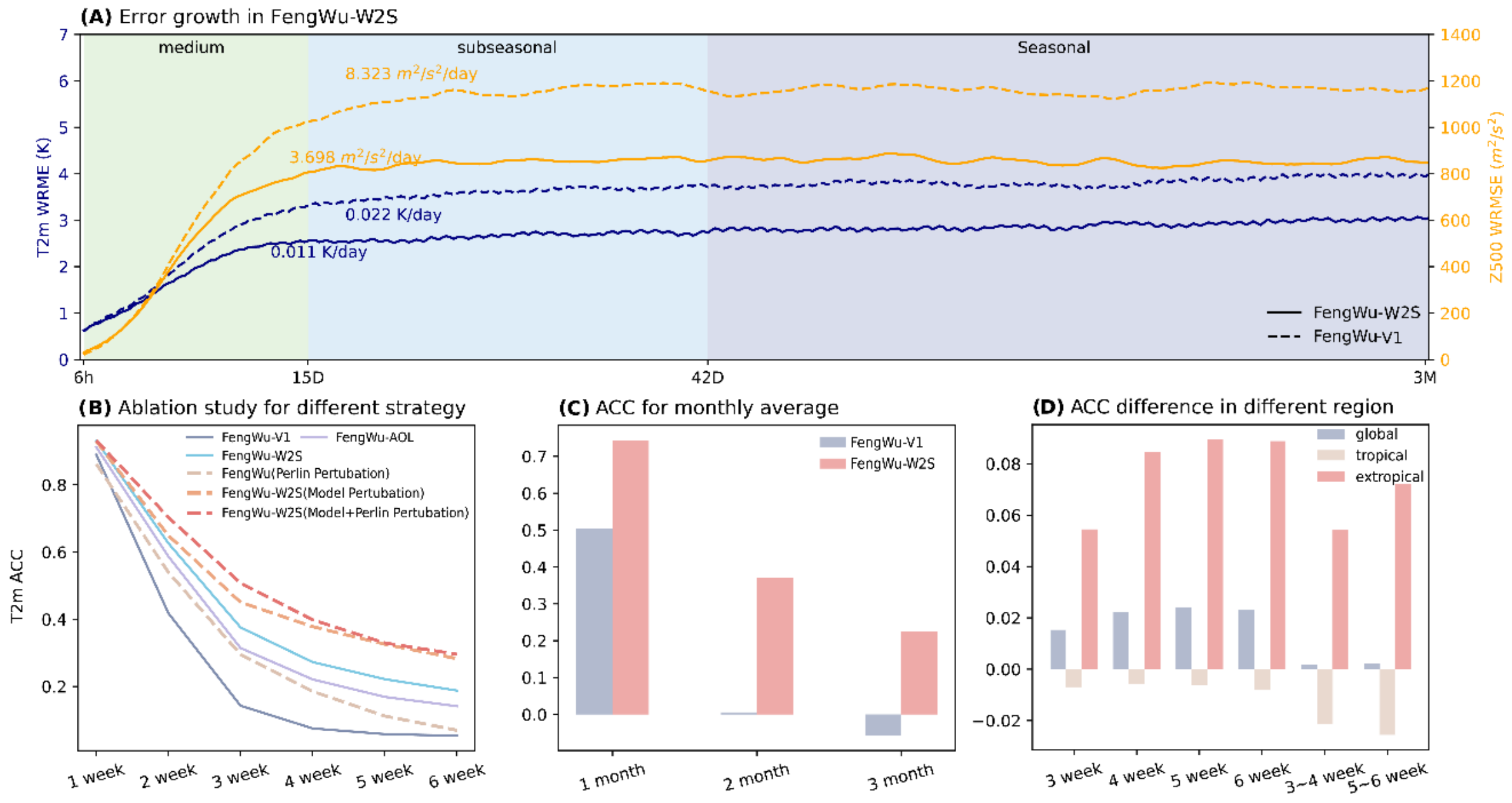

**Figure 5 The error growth and ablation study for FengWu-W2S.** (A) The error growth in AI model. The dashed (solid) lines represent the prediction of FengWu (FengWu-W2S), and the blue (orange) lines represent the prediction of T2m (Z500). (B) Ablation study of different strategies in FengWu-W2S$_o$, including FengWu-V1, FengWu-AOL with more variables but without energy and mass exchange modules, and FengWu-W2S with different ensemble strategies. (C) The forecast skill of monthly averaged T2m anomalies. (D) Differences in forecast skills across different regions between FengWu-W2S$_o$ and FengWu-W2S$_m$.

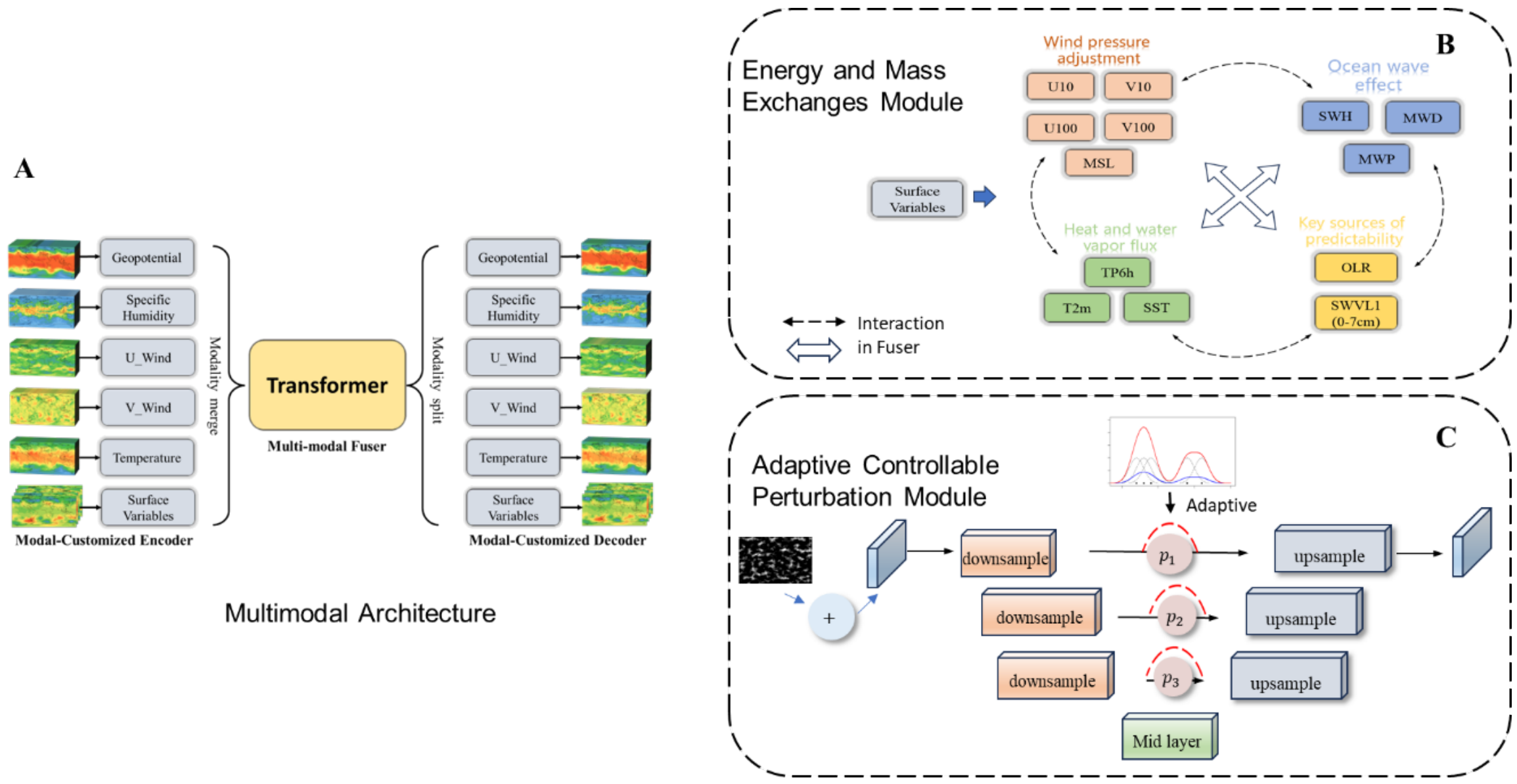

**Supplementary Figure 1. The main components of FengWu-W2S.** (A) The multimodal architecture of FengWu-W2S, which inherited from FengWu. (B) The surface data is divided into different modules for feature extract and facilitating the exchange of mass and energy between subsystem in the multi-modal fuser by a physically guided way. (C) Multi-level perturbation strategies for different variables that can be controlled manually.

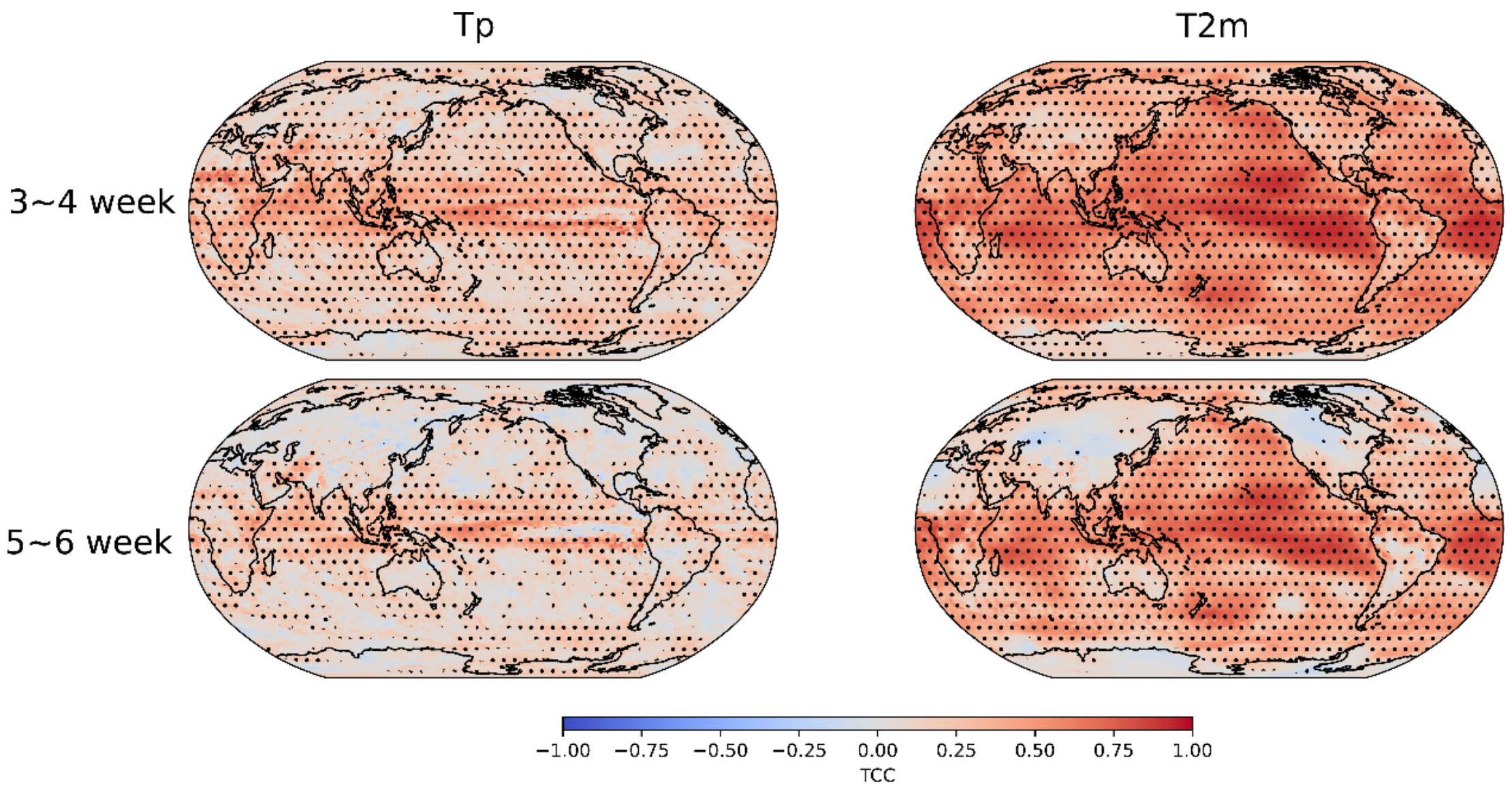

**Supplementary Figure 2. Spatial TCC skill distribution of FengWu-W2S with model's climatology.**

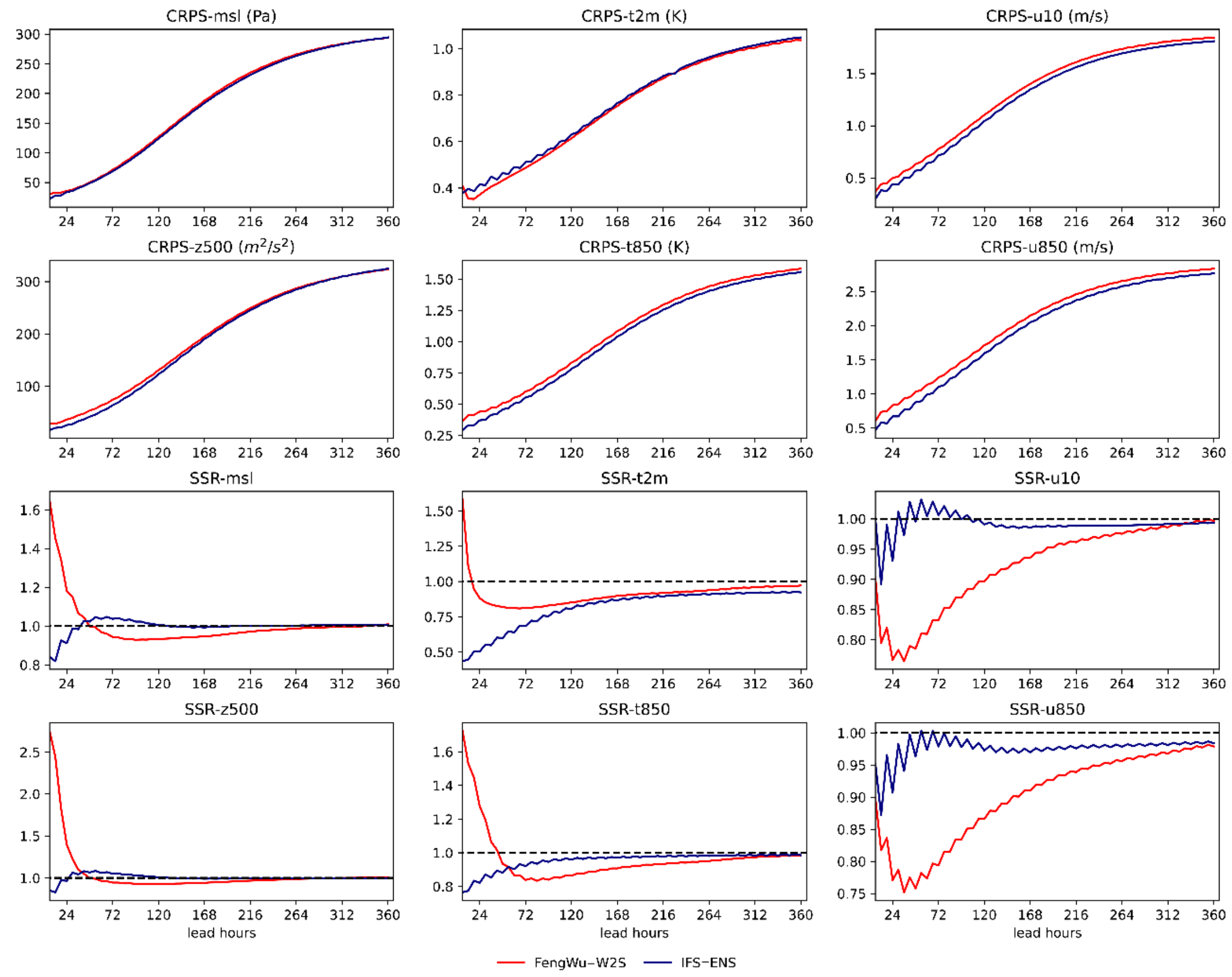

**Supplementary Figure 3. The medium-range weather forecast skill of FengWu-W2S.**

**Supplementary Table 1 A summary of all variable names and their abbreviations**

| Type | Long name | Abbreviation |
|---|---|---|
| Upper-air variable (13 levels 50, 100, 150, 200, 250, 300, 400, 500, 600, 700, 850, 925, and 1000 hPa) | Geopotential | Z |
| | Temperature | T |
| | Zonal wind | U |
| | Meridional wind | V |
| | Specific humidity | Q |
| Single-air variable | Temperature at 2 meters | T2m |
| | Total precipitation | TP |
| | Mean sea-level pressure | MSL |
| | Zonal wind at 10 meters | U10 |
| | Meridional wind at 10 meters | V10 |
| | Zonal wind at 100 meters | U100 |
| | Meridional wind at 100 meters | V100 |
| | Outgoing longwave radiation | OLR |
| Ocean variable | Sea surface temperature | SST |
| | Significant wave height | SWH |
| | Mean wave direction | MWD |
| | Mean wave period | MWP |
| Land variable | Water volume in the soil layer from 0 to 7 cm | SWVL1 |

Training period:1979-01-01~2016-12-31

Time resolution: 6 hourly

Spatial resolution:1.4° latitude/longitude